\RequirePackage{fix-cm}

\documentclass{svproc}      
\usepackage{graphicx}
\usepackage{hyperref}
\usepackage[vertfit]{breakurl} 
\usepackage{url}

\begin{document}
\mainmatter  
\title{An Integrated Approach for Video Captioning and Applications}


\titlerunning{An Integrated Approach for Video Captioning and Applications}       

\author{Soheyla Amirian\inst{1}\inst{2} \and Thiab R. Taha\inst{1} \and Khaled Rasheed\inst{1} \and Hamid R. Arabnia\inst{1}}
\authorrunning{Soheyla Amirian et al.} 

\institute{Department of Computer Science, \\The University of Georgia, Athens, GA 30602-7404, USA,\\
\and
\email{amirian@uga.edu}
}

\maketitle

\begin{abstract}

Physical computing infrastructure, data gathering, and algorithms have recently had significant advances to extract information from images and videos. The growth has been especially outstanding in image captioning and video captioning. However, most of the advancements in video captioning still take place in short videos.
In this research, we caption longer videos only by using the keyframes, which are a small subset of the total video frames. Instead of processing thousands of frames, only a few frames are processed depending on the number of keyframes. There is a trade-off between the computation of many frames and the speed of the captioning process. The approach in this research is to allow the user to specify the trade-off between execution time and accuracy. In addition, we argue that linking images, videos, and natural language offers many practical benefits and immediate practical applications. From the modeling perspective, instead of designing and staging explicit algorithms to process videos and generate captions in complex processing pipelines, our contribution lies in designing hybrid deep learning architectures to apply in long videos by captioning video keyframes. We consider the technology and the methodology that we have developed as steps toward the applications discussed in this research.

\begin{keywords}
Deep Learning, Image Captioning, Video Captioning, Keyframe.
\end{keywords} 
\end{abstract}

\section{Introduction}
\label{sec:intro}
\begin{figure}[htbp]
    \centering\includegraphics[width=0.5\textwidth]{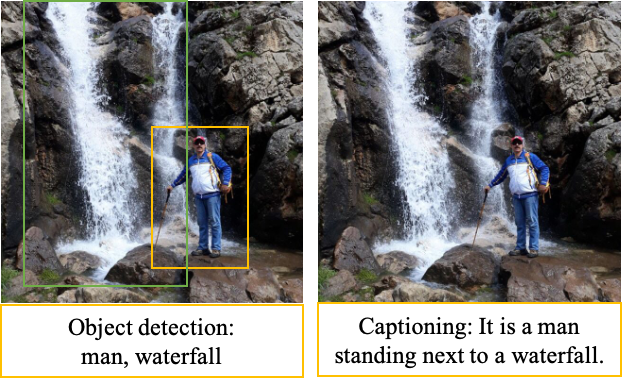}
    \caption{Left: detecting objects and information in an image. Right: detecting and describing the image}
\label{fig:chap1Ex1}

    \centering\includegraphics[width=\textwidth]{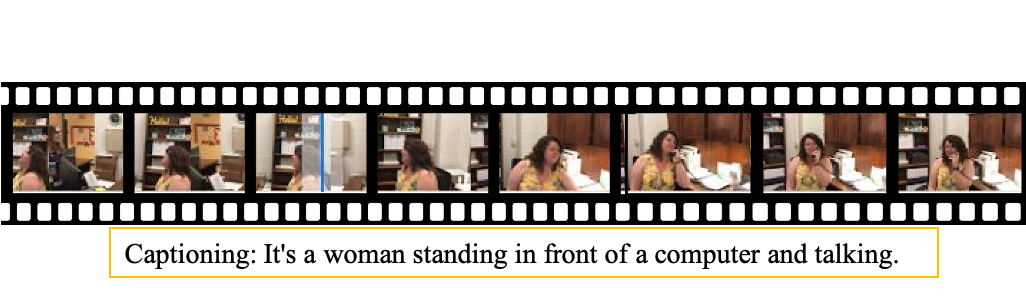}
    \caption{It describes the scenery and actions.}
\label{fig:chap1Ex2}
\end{figure}

Based on Wikipedia's definition, Artificial intelligence (AI) is intelligence demonstrated by machines, unlike the natural intelligence displayed by humans and animals, which involves consciousness and emotionality.
One of the researcher's goals in Artificial Intelligence and Computer Vision is to enable computers to understand the visual world around us. In addition, scientists try to communicate with machines in natural human language so that machines can benefit people by doing various tedious tasks. 
Humans can accomplish different tasks that need visual understanding, including communication and interpretation in natural language by looking at a picture. A human can extract the information and describe an immense amount of details. For instance, by looking at Figure \ref{fig:chap1Ex1}, we can immediately describe it as "It is a man standing on a rock next to a small waterfall. He has a blue-white outfit with a hat. He is holding something in his left hand.". In another example, Figure \ref{fig:chap1Ex2}, we can watch a video and describe it as "It is a woman, wearing a yellow top, standing in front of a computer and talking on the phone.".
Over the last few years, physical computing infrastructure, data gathering, and algorithms to extract information from images and videos have had significant and impressive advances. The growth has been especially outstanding in image captioning and video captioning. However, despite significant encouraging progress, most of the advancements in video captioning still take place in short videos.
In this research, we caption videos only by using the keyframes, which are a small subset of the total
video frames, to caption long videos. We apply this method to caption videos for real-world purposes.

Section \ref{sec:related} provides relevant deep learning structures and models \cite{amirian2018}.
In Section \ref{sec:challenge}, we analyze the challenges of extracting information from images and videos.
In Section \ref{sec:exp} we explain the process of generating descriptions by keyframes in long videos. Also we discuss the motivations of the proposal.
Section \ref{sec:cs} gives two case studies. One proposes an architecture that utilizes image/video captioning methods and Natural Language Processing systems to create a title and a concise abstract for a video. Then, we propose another application for video captioning for fostering and facilitating physical activities.

\section{Definition and Related Work}
\label{sec:related}
Deep Learning is a platform for solving impactful and challenging problems. Deep Learning allows computers to learn from experience and understand the world in terms of a hierarchy of concepts \cite{goodfellow2016deep}.In recent years, deep learning-based convolutional neural networks have positively impacted image recognition and increased flexibility. ...

\subsubsection{Image and Video Captioning}
Image and video captioning with deep learning are used for the difficult task of recognizing the objects and actions in an image or video and creating a concise meaningful sentence based on the contents found.
Many impressive studies have been done about image captioning~\cite{farhadi2010every,you2016image,wu2016value,karpathy2014deep,soh2019image}. In image captioning, Aneja et al. \cite{aneja2018convolutional} developed a convolutional image captioning technique with existing Long Short Term Memory (LSTM) techniques and analyzed the differences between Recurrent Neural Networks (RNN) based learning and their method. Their techniques differ primarily in their intermediary components. 
Aneja et al. use masked convolution in a CNN-based approach, whereas RNN employs LSTM or GRU. The intermediary component of Aneja et al. is feed-forward without any recurrent function, and their CNN with attention (Attn) achieved comparable performance to the RNN-based approach. They also experimented with an attention mechanism and attention parameters using the conv-layer activations. The results of the CNN+Attn method were increased relative to the LSTM baseline. For better performance on the MS COCO, they used ResNet features, and the results show that ResNet boosts their performance on the MS COCO. The results on MS COCO with ResNet101 and ResNet152 were impressive.

Video description is the automatic generation of meaningful sentences that describes the events in a video. Many researchers present different models on video captioning~\cite{donahue2015long,venugopalan2016improving,torabi2015using,krishna2017dense,wang2018video}, mostly with limited success and many constraints.
In video captioning \cite{soh9281287}, Krishna et al. \cite{krishna2017dense}, however, presented Dense-captioning, which focuses on detecting multiple events that occur in a video by jointly localizing temporal proposals of interest and then describing each with natural language. This model introduced a new captioning module that uses contextual information from past and future events to describe all events jointly. They implemented the model on the ActivityNet Captions dataset. The captions that came out of ActivityNet shifted sentence descriptions from being object-centric in images to action-centric in videos.
Park et al. \cite{sung2019adversarial} applied Adversarial Networks in their framework. They propose to use adversarial techniques during inference, designing a discriminator which encourages multi-sentence video description. They decouple a discriminator to evaluate visual relevance to the video, language diversity and fluency, and coherence across sentences on the ActivityNet Captions dataset.

\subsubsection{Sources of Data:}
The considerable majority of modeling approaches in this area fall under data-driven techniques, in which a model learns from human explanation data. 
It is therefore essential to highlight the available datasets.
There are a few datasets that relate the visual domain with the field of natural language.
This research uses the Flickr8K \cite{karpathy2014deep}, Flickr30K \cite{xu2015show,karpathy2014deep}, and Microsoft COCO \cite{xu2015show,chen2015microsoft} datasets consisting of a set of images, each annotated with five descriptions written by humans on the Amazon Mechanical Turk for image captioning models. We use a few different datasets for video captioning, and classify them into five domains based on the video contents: People, Open Subjects, Social Media, Cooking, and Movie \cite{}. 
For the experiments, we used videos from from the  Microsoft Video Description dataset (MSVD) and MSR Video to Text (MSR-VTT) Dataset.
The Microsoft Video Description dataset (MSVD) \cite{chen2011collecting} contains 1,970 YouTube clips with human-annotated sentences. The duration of each video in the MSVD dataset is typically between 10 to 25 seconds. On average, 41 descriptions for each video are there. In total, this video dataset has approximately 80,000 description pairs and about 16,000 vocabulary words, which Microsoft Research provides.
MSR Video to Text (MSR-VTT) \cite{xu2016msr} is created by collecting 257 popular queries from a commercial video search engine, with 118 videos for each query. MSR-VTT provides 41.2 hours of 10K web video clips with 200K clip-sentence pairs in total, covering a list of 20 categories.
A caption could be a sentence that may explicitly mention non-obvious aspects of the scene (such as people, locations, or dates) that the model cannot derive from the image alone (i.e., they add information). In the case of image-sentence datasets, the human annotators describe the image's content with a sentence. If the captions are inconsistently defined, it is impossible to achieve high accuracy.
Therefore, these datasets contain numerous image descriptions since it is challenging for people unfamiliar with the specific context to caption the image appropriately. 


\section{Challenges}
\label{sec:challenge}
It may seem effortless for humans to see scenes and describe them. However, this is a complicated task for computers. The process through which computers can gain high-level understanding from digital images or videos to understand and automate tasks that the human visual system can do is dense.
In the digital world, each image or video frame represents an extensive array of numbers called pixels. For example, each pixel indicates the brightness at any position. A typical image includes a few million of these pixels, and a computer must transform these values into semantic concepts. These concepts are categorized into different classes —for instance, humans, animals, objects, actions, and many others.
Moreover, a different type of object seen under different lighting conditions, with a different camera angle, or a different pose might depict another caption. Here, "a woman standing in front of a computer", however, the brightness values could change and result in a different caption.
Furthermore, patterns with very similar low-level statistics (high-frequency patterns) might instead be part of many different objects (planes, cars, and many others) or animals (cats, dogs, bears, and many others). Thus, the challenges are no less severe on the language side. 

A natural language description such as "It's a woman standing in front of a computer and talking" will be represented in the computer as a sequence of integers indicating the index of each word in a vocabulary, for example [2, 1803, 409, and many others].
All told, the first step of accurately identifying and naming different parts of an image requires a complex pattern recognition process that needs much computation.
Moreover, image captions often require detecting and describing complex high-level concepts that are not only visual but require problematic inferences such as actions. It requires detecting multiple objects and analyzing their poses, spatial arrangements, or even facial features for characters. For example, someone could be described as “delivering” something, “dancing,” or “jumping”. 
Moving on to videos, complexity increases as each video consists of many images (frames). In addition, time matters, so the sequence and order of frames would be important too. Consequently, there would be a massive computation process to extract all the information, and assign a meaningful caption to the video to describe it appropriately.

There has been reasonably good progress in deep learning, visual recognition, image captioning, and video captioning systems. 
Notably, the state-of-the-art image recognition models based on deep convolutional neural networks \cite{NIPS2012_4824} have become capable of distinguishing thousands of visual categories. Similarly, advances in related tasks such as segmentation and object detection have been dramatic \cite{redmon2018yolov3}. 
Much impactful research has been done on image captioning \cite{farhadi2010every,karpathy2014deep,wu2016value,you2016image}, and video captioning \cite{krishna2017dense,mun2019streamlined}.
In addition, many fundamental training datasets have become available. Together, these advances enabled many real-world applications. They include face detection, visual recognition, activity detection, photo search, automatic caption generation for images and videos for people who suffer from various degrees of visual impairment, general-purpose robot vision systems, and many others. Moreover, these advances can positively and significantly impact many other task-specific applications.

\begin{figure}[htbp]
    \centering\includegraphics[width=0.7\textwidth]{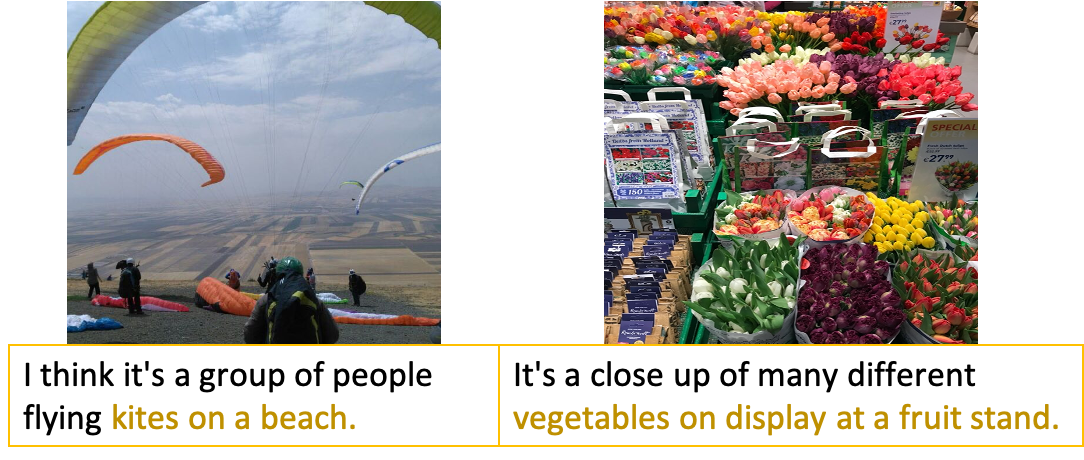}
    \caption{Here are two examples of inaccurate captioning.}
\label{fig:chap1flower}
\end{figure}

The overall approach in most of these applications is to model the visual recognition part. The first step is to extract the critical information to classify and detect objects, actions, and many others, from images. And then, the model attempts to caption images in real-time. 
For example, R-CNN significantly improves the quality of candidate bounding boxes and takes a deep architecture to extract high-level features \cite{girshick2014rich}. 
Fast and Faster R-CNN focus on speeding up the R-CNN framework by sharing computation and using neural networks to propose regions. Nonetheless, while Fast and Faster R-CNN offer speed and accuracy improvements over R-CNN, both still fall short of real-time performance \cite{Redmon_2016_CVPR}. On the other hand, deep Residual Networks (ResNet) present a residual learning framework and gain accuracy from considerably increased depth \cite{he2016deep}.
Temporal Action Detection is a tool that focuses on localizing the temporal extent of each action for a detected object \cite{caba2016fast}. 
Object detection benchmarks use a different set of chosen categories (e.g., “car,” “person,” “plant,” and many others). Different datasets for scene classification, action classification, or attribute classification have their own sets of categories. 
While visual classes constitute a convenient modeling assumption, this approach sometimes pales compared to the complexity of descriptions that humans can compose for images and videos (e.g., Figure \ref{fig:chap1flower}).

\section{Methodology and Experiments}
\label{sec:exp}

Each video consists of many frames. Some of these frames have the same contents, and some have essential information. Instead of giving all these frames to a video captioning model to assign a caption, we extract the video frames that contain necessary information called keyframes. Then we apply captioning techniques to those keyframes. Therefore, we caption videos only by using the keyframes, which are a small subset of the total video frames. Instead of processing tens of thousands of frames, a few frames are processed depending on the number of keyframes. 

\begin{figure}[htbp]
    \centering
    \includegraphics[width=\linewidth]{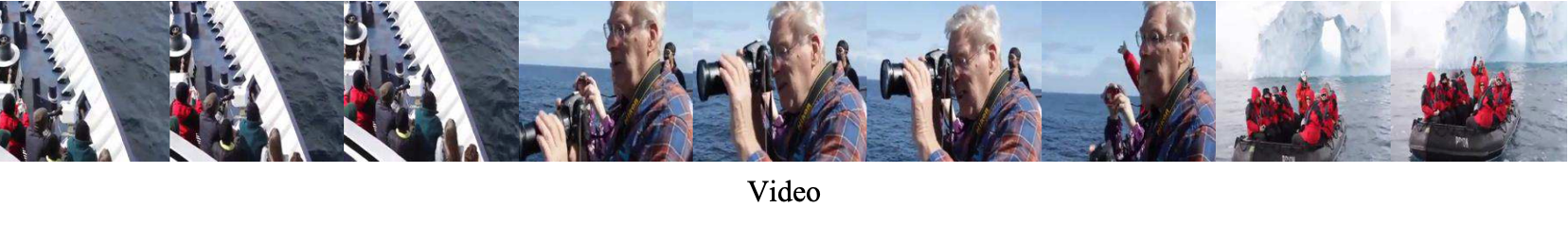}
    \caption{SA-LSTM: a man is in the water; RecNet-local: a man is taking pictures on boat; RecNet-global: people are riding a boat \cite{wang2018reconstruction}}
    \label{fig:1wang}
\end{figure}

Figure \ref{fig:1wang} is an example from  Wang et al. \cite{wang2018reconstruction} for a video captioning model that generates a caption based on the video. Wang et al. proposed a reconstruction network (RecNet) with an encoder-decoder-reconstructor architecture, which leverages both the forward (video to sentence) and backward (sentence to video) flows for video captioning. 

\begin{figure}[htbp]
    \centering
    \includegraphics[scale=0.6]{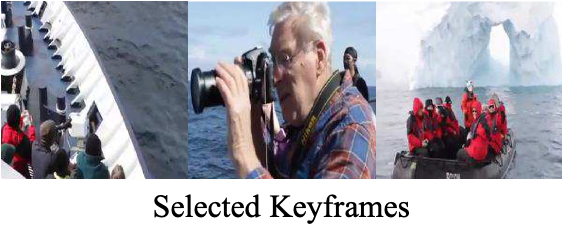}
    \caption{A group of people standing next to each other. A man is holding his camera up to take a picture. A group of people riding a boat across a body of water. \cite{luo2018discriminability}\\
    Ground Truth: bunch of people taking pictures from the boat and going towards ice.}
    \label{fig:1Luo}
\end{figure}

Next, Figure \ref{fig:1Luo} shows some selected keyframes from the sample video. Then, captions are generated by an image captioning model \cite{luo2018discriminability} model. 
It concludes with exciting results as we can see more descriptions and more details about the video.

The above contributions offer helpful and practical applications in the short term and will lead to enhanced machine understanding in the long term.

\subsubsection{Motivations} 
The present research is a step towards a future where we can interact with computers, incredibly where these interactions can be appropriately grounded in physical environments. Therefore, working towards artificially intelligent assistants will require us to make large amounts of information about how our world works available to computers.
Concretely, there are extensive sources of knowledge for this motivation. The physical domain contains information about the world, scenes, objects, and interactions (vision, language, and visual sensors). The Internet's digital domain includes a vast amount of information not accessible from the physical domain (e.g., what happened in 1981).  
Because vision and language are the primary means to access the world's knowledge, we must develop techniques that can relate information across these two areas instead of processing each one independently.
A solid short-term example might allow a computer to generate a caption based on the objects that it extracts from the images or videos. For instance, "it's a man that is standing in the snow","it's a wooden statue in a park".
In the longer-term future, the computer could understand more by examining the procedure and events in the images or video and inform us about it if my kid "wants to touch a pan on a stove" or if they "want to jump over a fire". Furthermore, the system we developed would open up doors for Generative Adversarial Networks (GANs) to be used for generating videos; such a thing does not presently exist. However, this is a step toward that goal.

\begin{figure}[htbp]
    \centering\includegraphics[width=\linewidth]{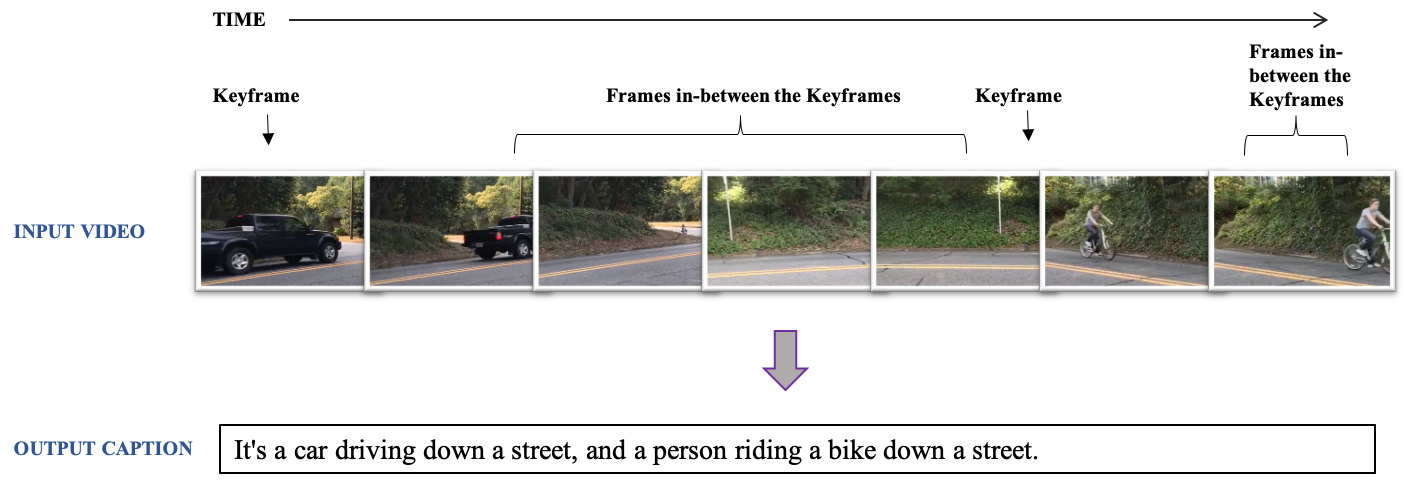}
    \caption{Notice the keyframes.}
\label{fig:chap1Ex3}
\end{figure}

The aspirations to select keyframes to generate captions and descriptions for a video can also be motivated by more concrete, short-term, and practical arguments. An Artificial Intelligence system reads a video; representative image frames are identified and selected. Indeed, selecting keyframes and information from all the video frames is essential and challenging.
Then, it captions the image frames. We use image captioning models that accelerate the processing of a video. Natural Language Processing generates captions, and text summarization generates a video title.  
Figure \ref{fig:chap1Ex3} shows an example for selecting video keyframes and captioning them by image captioning model. Here, it captioned the video as "It's a car driving down a street, and a person riding a bike down a street." by selecting the keyframes. 
We demonstrate that our models, the procedures they take advantage of, and the interactions they enable, are a path towards Artificial Intelligence. In addition, we argue that linking images, videos, and natural language offers many practical benefits and immediate valuable applications. From the modeling perspective, instead of designing and staging explicit algorithms to process videos and generate captions in complex processing pipelines, our contribution lies in designing hybrid deep learning architectures to apply in long videos by captioning video keyframes. In addition, we show how our model facilitates tedious human tasks by utilizing different deep learning models. 
\subsubsection{Challenges of this Approach}
However, the Keyframes approach also poses some challenges. For example, which frames should we call keyframes? And how can we select the keyframes for captioning? One common criticism is that we may miss some vital information when we choose only some keyframes in a video. In addition, selecting the keyframes may generate descriptions that may not be accurate. Further, assessing the accuracy itself is complex. Presently, we have to evaluate accuracy to a human interpretation. 
In our work, we select each keyframe based on a transition that happens in the video. And we use the state-of-the-art image captioning model that offers the highest correlations with human judgments. We expect further improvement by evaluating the whole end caption with the integration model in the future.

\begin{figure}[htbp]    
\centering
    \includegraphics[width=\linewidth]{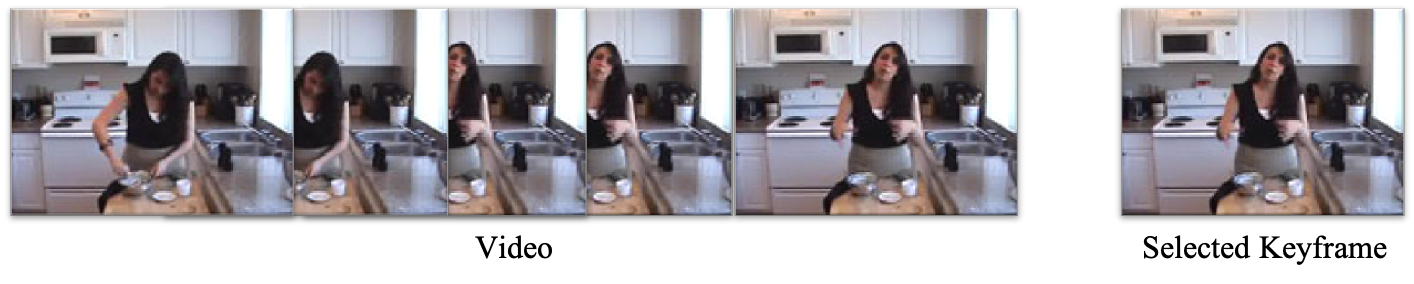}
    \caption{Left: a-LSTMs: a woman is washing her hands \cite{gao2017video};
    Right: a woman standing in a kitchen next to a stove top oven \cite{luo2018discriminability}. GT: A woman is stirring some ingredients.}
    \label{fig:2Gao_woman}
\end{figure}

Another challenge is that this approach integrates the keyframe selection (which images give us more information), visual recognition task (how to extract the information from keyframes), with the language modeling task (how to generate a meaningful caption based on the information we have). Therefore, it may feel safer to disintegrate these tasks, study them separately and then compose them to form the full model later. 
On the other hand, addressing these tasks combined will allow us to formulate a single model that automatically generates captions for a video.
In Figure \ref{fig:2Gao_woman}, we picked an example from Gao et al. \cite{gao2017video}. First, we chose a keyframe from the video. Then we generated captions with an image captioning model \cite{luo2018discriminability}. The result is notably impressive.

\begin{figure}[htbp]
\centering
    \includegraphics[width=\linewidth]{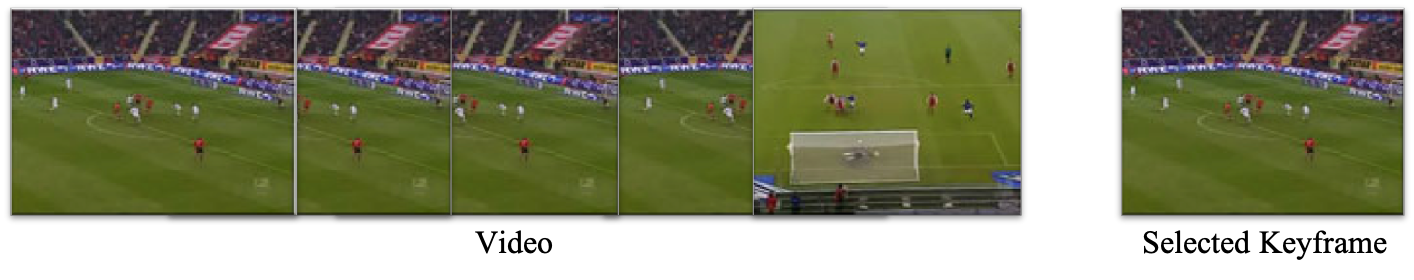}
    \caption{Left: a-LSTMs: a group of man are playing football \cite{gao2017video};
    Right: A blurry image of a baseball game in progress. A baseball game is being played on a green field.  \cite{luo2018discriminability}. GT:People are playing football.}
    \label{fig:2Gao_football}
\end{figure}

\begin{figure}[htbp]
\centering
    \includegraphics[scale=0.6]{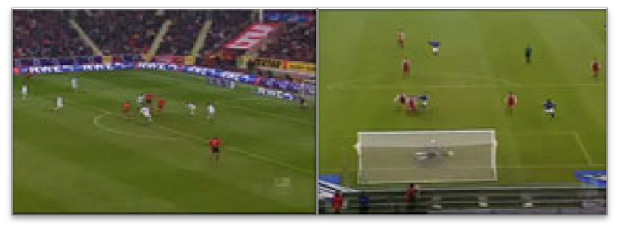}
    \caption{A blurry image of a baseball game in progress. A baseball game is being played on a green field.  \cite{luo2018discriminability}.
    }
    \label{fig:2Gao_foot2}
\end{figure}

In another example, Figure \ref{fig:2Gao_football}, we see that one keyframe does not give us a clear caption. Therefore, we need to choose more keyframes from the video to have a better caption. Consequently, we added another keyframe. Figure \ref{fig:2Gao_foot2} shows the generated caption.

As a result, common sense tells us that the more frames we process in a video, the more accurate a caption will be. However, more computations increase the expense. Therefore, there is a trade-off between the computation of many frames and the speed of the captioning process. However, the approach in this research is to allow the user to specify the trade-off between execution time and accuracy.

\section{Case Studies}
\label{sec:cs}

In an application, we applied our approach in keyframes to make titles for videos \cite{title2020}. The proposed application introduces an architecture that utilizes image/video captioning methods and Natural Language Processing systems to create a title and a concise abstract for a video. If we can generate reasonably meaningful titles for videos, the search engines could use them as metadata. For example, we search for videos that contain a woman with a red dress and sunglasses. Generating titles would significantly help search engines to search for videos. Other application domains include the cinema industry, security surveillance, video databases/warehouses, data centers, and many others, can potentially utilize such a system.

\begin{figure}[htbp]
    \centering
    \includegraphics[width=\linewidth]{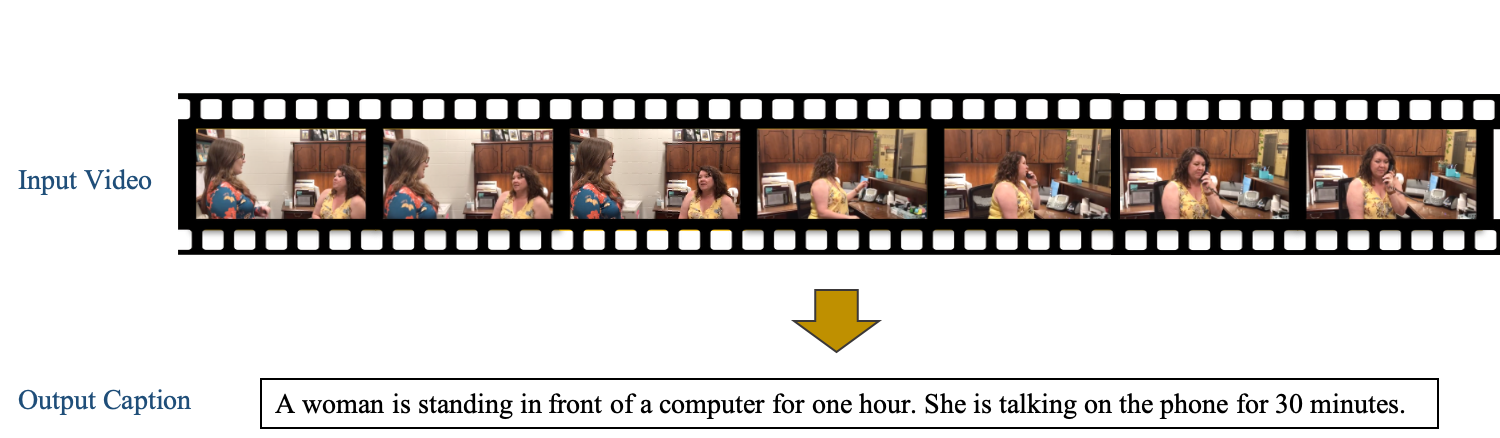}
    \caption{As an example of the proposed framework, the generated story describes the type of each physical activity and the corresponding duration of the video.}
    \label{fig:1physicalAct}
\end{figure}

Next, a novel application \cite{amirian2020fostering} involves taking in a video and directly generating captions describing a person's activities during a period without being constrained to time or location. The proposed model could be assistive technology to foster and facilitate physical activities. This framework could potentially help people trace their daily movements to reduce the health risks of an inactive lifestyle by managing their activities. Our work could be a healthcare application used by physicians or the public. Figure \ref{fig:1physicalAct} illustrates this model. 

\section{Discussion and Future Work}
\label{sec:conclusion}

We have observed active and fast advances in the field of Computer Vision (deep learning, image, and video captioning) over the last few years. Progress in this area has unlocked a wide variety of ambitious problems that once defied our efforts.
In particular, in this research, we developed different frameworks and techniques that push the frontier of video captioning. By selecting the keyframes in a video and proposing frameworks for various applications, we step towards Artificial Intelligence agents that can perceive the visual world and interact with us naturally. 
We argued that by selecting keyframes from videos, we could assign captions to the long videos and use them in different applications, accelerating the video captioning process. We evaluated this captioning qualitatively in this work; quantitative assessment could be the subject of future studies.

Despite recent rapid progress in image/video captioning, it is clear that many challenges remain in the vision of machines that can sense the visual world and interact with us naturally to utilize them in many applications. We can employ the captioning process in search engines, the cinema industry, and computer vision tasks. Notably, we pick some of the keyframes in the video. If the keyframes do not give us a good caption, we can add more. There is a trade-off between execution time for video captioning and the accuracy of the generated captions.
In other words, our goal is to reduce the computation process and speed up the captioning task for videos. Even if the generated caption for the video is not accurate, we can enhance it further by using crowdsourcing.
We hope that our ideas can be reused and help inspire future work.

\begin{figure}[htbp]
\centering
    \includegraphics[scale=0.5]{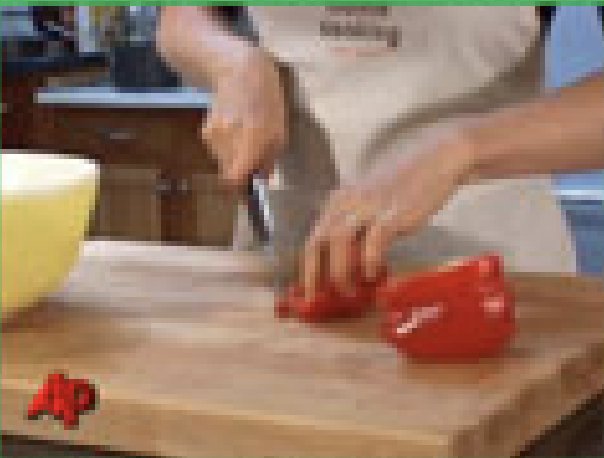}
    \caption{Video: a-LSTMs: a woman is slicing a tomato into pieces \cite{gao2017video};
    Only one frame: a person standing at a counter with a ball. \cite{luo2018discriminability}\\ 
    GT:a woman is slicing a red pepper.}
    \label{fig:Gao_pepper}
\end{figure}

The remaining challenges are best illustrated with an example. Consider the image in Figure \ref{fig:Gao_pepper}. As you can see, we need more frames to extract the pepper and by having only a frame we may not have an exact caption. However, we proved that the keyframes are a good method to caption for long videos. In addition, there is a trade off between execution time for video captioning and the accuracy of the captions. Even if the video caption is not accurate, we can enhance it further by using crowdsourcing. Also, even if the keyframe don’t give us a good result, it’s not a big deal. We can add more keyframes like the example we had here.
Experiments demonstrated that our method using keyframes can achieve competitive results for video captioning.


As a solid short-term example, this might allow a computer to generate a caption based on the objects that it extracts from the images or videos. 
The present research is a step towards a future where we can interact with computers
In the longer-term future, the computer could understand more by examining the procedure and events in the images or video and inform us about it if my kid "wants to touch a pan on a stove" or if they "want to jump over a fire.". 
Consequently, the system we developed would open up doors for Generative Adversarial Networks (GANs) to be used for generating videos; such a thing does not presently exist. However, this is a step toward that goal.

\section{Acknowledgement}
\label{sec:ack}
We gratefully acknowledge the support of NVIDIA Corporation with the donation of the Titan V GPU used for this research.

\bibliographystyle{spmpsci}
\bibliography{reference}

\end{document}